\PassOptionsToPackage{table}{xcolor}

\documentclass[]{mpai_report}

\usepackage[toc,page,header]{appendix}

\usepackage{natbib}

\usepackage{CJKutf8}

\usepackage{xargs}  

\usepackage{todonotes}  

\usepackage{multirow}

\usepackage{cleveref}

\usepackage{amsmath}
\usepackage{dsfont}

\usepackage{subcaption}

\usepackage{svg}

\usepackage{mathrsfs}
\usepackage{adjustbox}
\usepackage{multirow}
\usepackage{multicol}
\usepackage{tcolorbox}
\usepackage{changepage}
\usepackage{graphicx}
\usepackage{amssymb}
\usepackage{array}
\usepackage{bm}
\usepackage{hyperref}

\usepackage{minitoc}
\usepackage{pifont}

\definecolor{categorygreen}{RGB}{232, 245, 233}
\definecolor{categoryorange}{RGB}{255, 243, 224}
\definecolor{categorygray}{RGB}{245, 245, 245}
\definecolor{categorypurple}{RGB}{243, 229, 245}
\definecolor{oursblue}{RGB}{232, 240, 254}

\newtcbox{\hlprimarytab}{on line, rounded corners, box align=base,
  colback=green!10, colframe=white, size=fbox, arc=3pt,
  before upper=\strut, top=-2pt, bottom=-4pt, left=-2pt, right=-2pt, boxrule=0pt}

\newtcbox{\hlsecondarytab}{on line, rounded corners, box align=base,
  colback=red!10, colframe=white, size=fbox, arc=3pt,
  before upper=\strut, top=-2pt, bottom=-4pt, left=-2pt, right=-2pt, boxrule=0pt}

\title{TouchWorld: A Predictive and Reactive Tactile Foundation Model for Dexterous Manipulation}
\renewcommand{\authorfont}{\fontsize{8.5}{10}\selectfont}

\author[1,2*]{Jianyi Zhou}
\author[1,2*]{Feiyang Hong}
\author[1,2*]{Yunhao Li}
\author[1,2]{Yicheng Zhao}
\author[1,2]{Yongjue Cen}
\author[1,2]{Zirui Liu}
\author[1,2]{Jiakang Huang}
\author[1,2]{Zirui Chen}
\author[1,2]{Ruiyang Zhang}
\author[1,2]{Weizhuo Zhu}
\author[1,2]{Xuhua Song}
\author[1,2\text{\ding{41}}]{Shuo Yang}

\renewcommand\affiliation[2][]{\addtolist[#1]{#2}{\affiliationlist}{\affiliationformat}{\\}}
\affiliation[1]{Harbin Institute of Technology, Shenzhen}
\affiliation[2]{PHANES AI}
\contribution[*]{Equal contribution}
\contribution[\text{\ding{41}}]{Corresponding author}

\abstract{
Dexterous manipulation in everyday environments requires both anticipation and reaction: a robot must predict how contact should evolve while rapidly correcting local errors caused by slip, misalignment, unstable grasping, or force mismatch. Vision and language provide semantic and geometric guidance, but they cannot reliably reveal hidden contact states such as force, slip, and contact stability. Although tactile sensing exposes these physical cues, most existing policies treat touch as a low-frequency observation stream within a monolithic action model, coupling slow task reasoning, action generation, and fast contact feedback in a single loop.
We introduce \textbf{TouchWorld}, a predictive-and-reactive tactile foundation model for dexterous manipulation. TouchWorld uses a hierarchical policy that separates vision-language subtask planning, tactile world-model prediction, visuo-tactile goal-conditioned action generation, and high-frequency tactile residual refinement. A \textbf{High-Level Planning Layer} produces executable subtasks and predicts tactile subgoals; a \textbf{Visuo-Tactile Goal-Conditioned Policy} generates nominal action chunks; and a \textbf{Tactile-Conditioned Refinement Policy} performs online residual correction using recent tactile and proprioceptive feedback. By using touch as both a predictive contact reference and a fast feedback signal, TouchWorld preserves the semantic generalization of vision-language-action policies while improving local contact adaptation. Across six long-horizon and contact-rich dexterous manipulation tasks, TouchWorld achieves 65.0\% success in the clean setting and 53.7\% success under human perturbations, outperforming the strongest baseline by 15.7 and 18.5 percentage points, respectively.

}
\date{\today}
\correspondence{\email{shuoyang@hit.edu.cn}}

\checkdata[Project Page]{https://phanes-lab.github.io/TouchWorld-website/}

\begin{document}
\begin{CJK*}{UTF8}{gbsn}

\maketitle

\section{Introduction}
Robots operating in everyday environments must perform manipulation tasks that go beyond reaching and moving objects.
Many daily skills, such as watering, power-plug insertion, cup insertion and tissue pulling, require the robot to anticipate how contact should evolve and to react when the actual contact deviates from expectation.
While vision and language provide semantic and geometric guidance, they often cannot reveal whether a grasp is stable, an object is slipping, an insertion is aligned, or the applied force is appropriate.
Tactile feedback provides direct access to these local contact states~\cite{zhao2025touchbeginsvisionends,huang2026spatiallyanchoredtactileawareness,zhang2025tavlaelucidatingdesignspace,yuan2026vtamvideotactileactionmodelscomplex,huang2025tactilevlaunlockingvisionlanguageactionmodels,zheng2026omnivtavisuotactileworldmodeling}, making it essential for robust contact-rich manipulation.
This naturally introduces a multi-timescale control problem.
Semantic task reasoning evolves slowly, visuo-tactile action generation operates at an intermediate action-chunk rate, and tactile feedback must support fast local correction when contact changes.

Despite recent progress in vision-language-action policies~\cite{black2026pi0visionlanguageactionflowmodel,intelligence2025pi05visionlanguageactionmodelopenworld,nvidia2025gr00tn1openfoundation,zheng2026egoscalescalingdexterousmanipulation,liu2025rdt1bdiffusionfoundationmodel,wang2026qwenvlaunifyingvisionlanguageactionmodeling,intelligence2026pi07steerablegeneralistrobotic,kim2024openvla,kim2025finetuningvisionlanguageactionmodelsoptimizing,dai2026conlacontrastivelatentaction,intelligence2025pi06vlalearnsexperience}, most existing systems still generate actions through a single monolithic model.
Tactile observations, when available, are often appended as additional input tokens and processed at the same rate as visual-language context~\cite{yuan2026ftp1generalistfoundationtactile}.
This design is poorly matched to contact-rich manipulation~\cite{xue2025reactivediffusionpolicyslowfast,xue2026tube}.
Slow semantic reasoning, intermediate action-chunk generation, and fast contact correction are forced to share the same model capacity and control loop~\cite{zhao2023learningfinegrainedbimanualmanipulation,chi2025diffusion}.
As a result, a policy may understand the task and produce plausible motions, yet still fail when stable grasping, force regulation, slip recovery, or precise insertion requires rapid tactile feedback.

We propose \textbf{TouchWorld}, a predictive-and-reactive tactile foundation model for dexterous manipulation.
TouchWorld is implemented as a hierarchical manipulation policy that uses touch in two complementary ways: a predictive pathway anticipates future contact-aware goals, and a reactive pathway corrects local execution errors online.
The system contains a multi-timescale hierarchy.
The \textbf{High-Level Planning Layer} runs at a slow semantic rate and contains the \textbf{Subtask Planner}, which produces executable subtasks, and the \textbf{Tactile World Model}, which predicts tactile subgoals for the current subtask.
A \textbf{Visuo-Tactile Goal-Conditioned Policy} runs at an intermediate action-chunk rate and generates nominal action chunks from the subtask, predicted tactile subgoals, visual observations, proprioception, and tactile observations.
A \textbf{Tactile-Conditioned Refinement Policy} refreshes local residual corrections inside the robot control loop using recent tactile and proprioceptive histories.

This design keeps semantic reasoning, predictive goal generation, and tactile feedback correction on separate time scales.
Within the High-Level Planning Layer, the Subtask Planner provides long-horizon structure and the Tactile World Model supplies future visual-tactile subgoals for contact-aware execution, while the refinement policy provides a fast feedback path when real contact deviates from the prediction.
Together, these components preserve the semantic generalization of vision-language-action policies while strengthening robustness in contact-rich dexterous manipulation.

\begin{figure}[t]
\centering
\includegraphics[width=\linewidth]{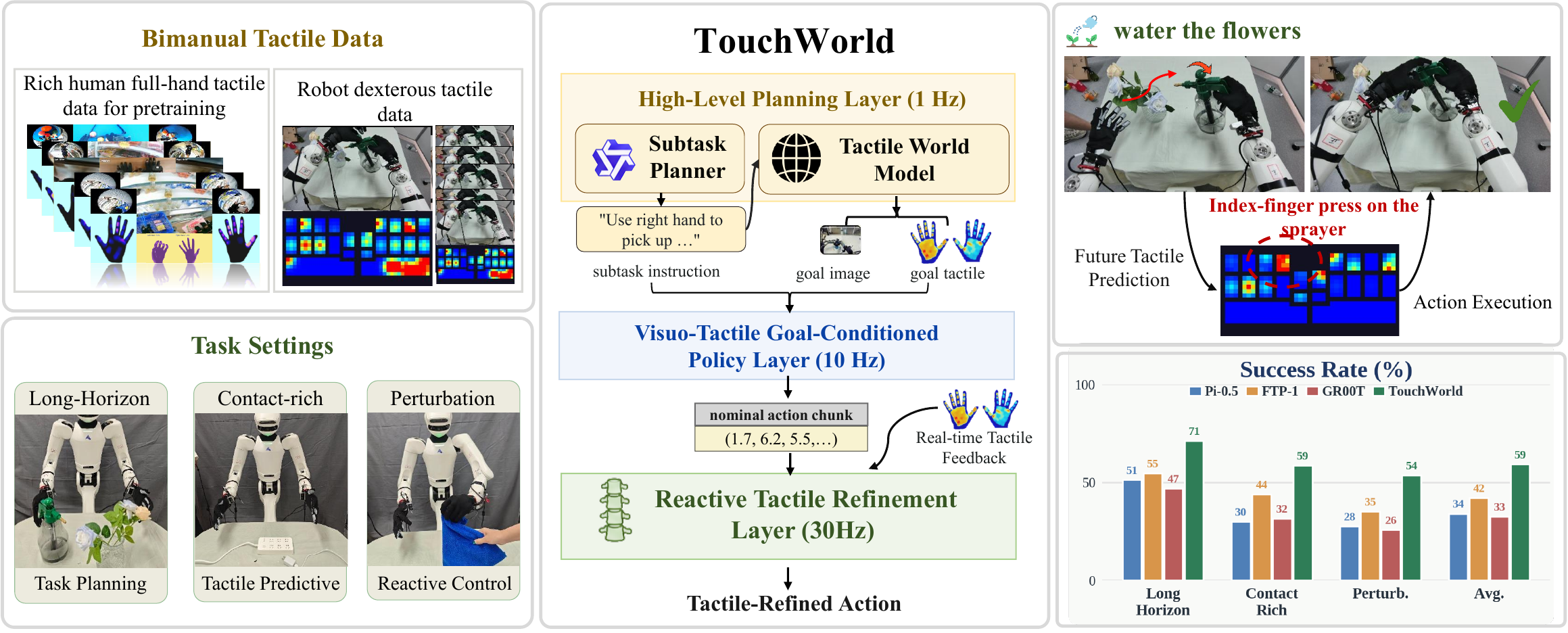}
\caption{Conceptual overview of TouchWorld. The High-Level Planning Layer contains a Subtask Planner that produces executable subtasks and a Tactile World Model that predicts visual-tactile subgoals. A visuo-tactile goal-conditioned policy generates nominal action chunks, and a tactile-conditioned refinement policy refines the final action online using high-frequency tactile feedback.}
\label{fig:teaser}
\end{figure}

The central question of this work is whether a tactile foundation policy can become more robust in contact-rich, long-horizon manipulation by explicitly separating semantic planning, tactile world-model prediction, visuo-tactile action generation, and tactile-conditioned feedback refinement.
To answer this question, we design TouchWorld around three capabilities:
semantic decomposition for long-horizon tasks, predictive tactile-goal conditioning for action generation, and reactive tactile correction under local contact perturbations.
We evaluate TouchWorld on six real-robot tasks and their human-perturbation variants: Water Flower, Tabletop Clearing, Cup Insertion, Power Plug Insertion, Pot Wiping, and Tissue Pulling.
TouchWorld achieves 65.0\% average success in the clean setting and 53.7\% under human perturbations, outperforming the strongest baseline by 15.7 and 18.5 percentage points, respectively.

This work makes three main contributions:

\begin{itemize}
    \item We introduce \textbf{TouchWorld}, the first predictive-and-reactive tactile foundation model for dexterous manipulation, implemented as a hierarchical manipulation policy that jointly supports tactile-goal prediction and online tactile feedback correction.
    
    \item We develop a \textbf{multi-timescale tactile policy architecture} that couples a slow High-Level Planning Layer, which contains the Subtask Planner and Tactile World Model, with visuo-tactile goal-conditioned action generation and high-frequency tactile residual refinement.
    
    \item We establish a six-task real-robot benchmark with clean and human-perturbation settings, where TouchWorld achieves 65.0\% average success in the clean setting and 53.7\% under human perturbations, outperforming the strongest baseline by 15.7 and 18.5 percentage points.
\end{itemize}

\section{Method}
\label{sec:method}

Given a task instruction $\ell$, current multi-view images $\mathcal{I}_t$, proprioceptive state $\mathbf{s}_t$, tactile observations $\mathcal{X}_t$, and high-level memory $m_t$, TouchWorld predicts the executed action $\mathbf{a}_t$ through the High-Level Planning Layer and two action policies.
The High-Level Planning Layer contains the Subtask Planner and Tactile World Model:

\begin{equation}
    \ell_t^{\mathrm{sub}} = \pi_{\mathrm{subtask}}(\ell, \mathcal{I}_t, m_t),
\end{equation}

\begin{equation}
    g_t = \pi_{\mathrm{world}}(\ell, \ell_t^{\mathrm{sub}}, \mathcal{I}_t, \mathcal{X}_t),
\end{equation}

\begin{equation}
    \left(\hat{\mathbf{A}}_{t:t+H-1}, \mathbf{c}_t\right) =
    \pi_{\mathrm{goal}}(\ell, \ell_t^{\mathrm{sub}}, g_t, \mathcal{I}_t, \mathbf{s}_t, \mathcal{X}_t),
\end{equation}

\begin{equation}
    \tilde{\mathbf{A}}_{\tau:\tau+W-1} =
    \pi_{\mathrm{tactile}}(\hat{\mathbf{A}}_{\tau:\tau+W-1}, \mathbf{s}_{\tau-k:\tau}, \mathcal{X}_{\tau-k:\tau}, \mathbf{c}_t),
\end{equation}

Here $\pi_{\mathrm{subtask}}$, $\pi_{\mathrm{world}}$, $\pi_{\mathrm{goal}}$, and $\pi_{\mathrm{tactile}}$ denote the Subtask Planner, Tactile World Model, Visuo-Tactile Goal-Conditioned Policy, and Tactile-Conditioned Refinement Policy, respectively.
The variable $\ell_t^{\mathrm{sub}}$ is the executable subtask at time $t$, and $g_t$ is the predicted visual-tactile subgoal for that subtask.
The memory $m_t$ stores previous subtasks, predicted tactile subgoals, and execution status.
The nominal policy outputs a nominal action chunk $\hat{\mathbf{A}}_{t:t+H-1}$ and context token $\mathbf{c}_t$, where $H$ is the nominal action horizon.
At refinement step $\tau$ inside the current chunk, the refinement policy uses a sliding nominal-action lookahead window of length $W$ together with recent proprioceptive and tactile histories, $\mathbf{s}_{\tau-k:\tau}$ and $\mathcal{X}_{\tau-k:\tau}$, where $k$ is the feedback-history length.
The controller executes the current action $\mathbf{a}_\tau$ from this corrected window during high-rate control.

\begin{figure}[t]
\centering
\includegraphics[width=\linewidth]{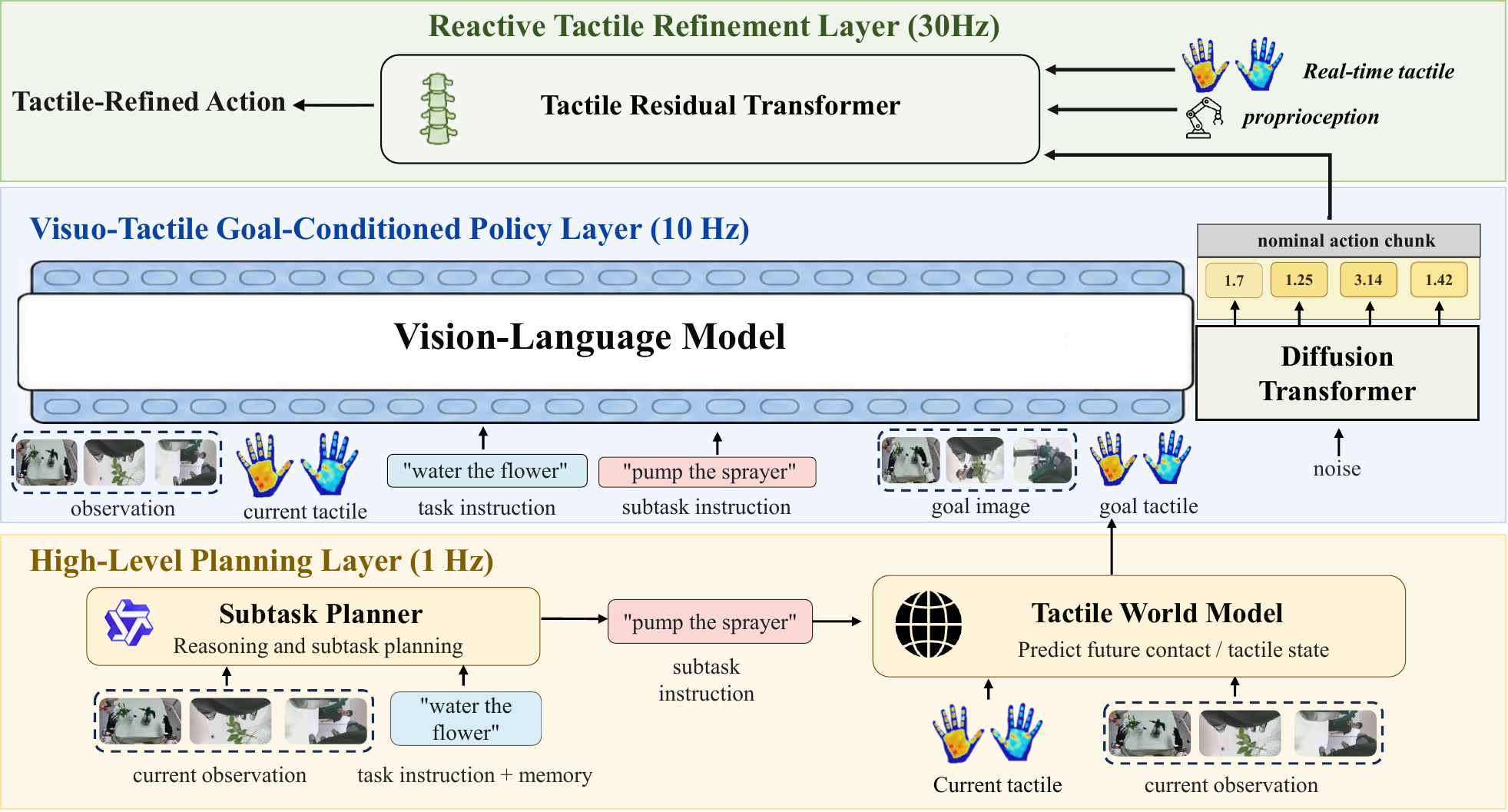}
\caption{TouchWorld architecture. The High-Level Planning Layer runs at a slow semantic rate and contains the Subtask Planner, which produces executable subtasks, and the Tactile World Model, which predicts visual-tactile subgoals. The Visuo-Tactile Goal-Conditioned Policy generates nominal action chunks at an intermediate rate, and the Tactile-Conditioned Refinement Policy refines execution inside the high-frequency robot control loop using tactile and proprioceptive feedback.}
\label{fig:architecture}
\end{figure}

\subsection{High-Level Planning Layer}
\label{sec:high_level_planning}

The High-Level Planning Layer performs slow semantic reasoning and predictive goal generation.
It contains the Subtask Planner and the Tactile World Model, both operating on the high-level memory $m_t$.

\noindent\textbf{Subtask Planner.}
The Subtask Planner receives the task instruction, current visual observations, and high-level memory, then maintains a compact task state for downstream control.
Rather than producing low-level commands, it updates the semantic phase of the task at a slow rate and emits an executable subtask for downstream policy conditioning.
The memory $m_t$ carries previous subtasks, predicted subgoals, and execution status, allowing the Subtask Planner to reason over task progress while using only the current observation at each update.

We represent the structured Subtask Planner output as:

\begin{equation}
    o_t^{\mathrm{sub}} = \{\ell, \ell_t^{\mathrm{sub}}, r_t\},
\end{equation}

where $o_t^{\mathrm{sub}}$ is the structured planner output, $\ell$ is the original task instruction, $\ell_t^{\mathrm{sub}}$ is the executable subtask, and $r_t$ is optional free-form reasoning.
The original task instruction and executable subtask are both exposed to the lower-level VLA policy, while free-form reasoning is kept outside the action interface.

\noindent\textbf{Tactile World Model.}
The Tactile World Model provides the predictive component of the High-Level Planning Layer.
Conditioned on the current observation and high-level state, it predicts future visual-tactile subgoals that describe the expected contact outcome of the current subtask.
These predictions are updated only when the Subtask Planner detects a new subtask or a meaningful task-state change, so stable task phases reuse the previous tactile subgoal instead of repeatedly regenerating future predictions.
The predicted subgoals serve as contact-aware references for the Visuo-Tactile Goal-Conditioned Policy.

\subsection{Visuo-Tactile Goal-Conditioned Policy}
\label{sec:goal_conditioned_policy}

The Visuo-Tactile Goal-Conditioned Policy is a tactile diffusion Transformer policy.
It receives the current subtask, visual observations, proprioception, and tactile observations, and predicts a nominal action chunk.
For the nominal VLA policy, tactile observations are converted into a unified image representation and processed together with visual observations by the vision-language branch.
A diffusion Transformer action expert then performs flow-matching action generation conditioned on the visual, tactile-image, language, and proprioceptive context.
When Tactile World Model predictions are available, predicted subgoal images or tactile subgoal latents can be appended as additional goal context; otherwise the policy conditions on the current observation and current subtask alone.

Let $\ell_{\mathrm{policy}}$ denote the prompt passed into the visuo-tactile policy.
We compose it from both the original task instruction and the current executable subtask:

\begin{equation}
    \ell_{\mathrm{policy}} =
    \texttt{Task: } \ell
    \oplus
    \texttt{ Current subtask: } \ell_t^{\mathrm{sub}}.
\end{equation}

Here $\oplus$ denotes string concatenation.
This prompt preserves the long-horizon task context while keeping the current control target explicit.

\noindent\textbf{Tactile image representation.}
The nominal policy uses a single image-form tactile interface.
Raw tactile readings are rendered or normalized into tactile images, so the VLA backbone receives touch in the same representational form as visual observations.
This design keeps the nominal action policy compatible with image-language pretraining and avoids introducing modality-specific tactile encoders into the VLA branch.

\noindent\textbf{Action generation.}
The policy predicts an action chunk $\hat{\mathbf{A}}_{t:t+H-1}$ using flow matching.
During training, noisy actions are sampled from the interpolation between data actions and Gaussian noise.
During inference, the model starts from Gaussian noise and integrates the predicted velocity field to produce a nominal action sequence.

\subsection{Tactile-Conditioned Refinement Policy}
\label{sec:tactile_refinement_policy}

The Tactile-Conditioned Refinement Policy converts the nominal action chunk into a locally corrected action window.
It operates at a faster feedback rate than the Visuo-Tactile Goal-Conditioned Policy so that contact changes can influence control without waiting for the next nominal chunk.
We implement this policy with a \textbf{Tactile Residual Transformer (TRT)}.
At refinement step $\tau$, TRT takes a sliding nominal-action lookahead window, recent proprioceptive and tactile histories, and the policy context token, and predicts a residual action window:

\begin{equation}
    \Delta \mathbf{A}_{\tau:\tau+W-1} =
    f_{\phi}(\hat{\mathbf{A}}_{\tau:\tau+W-1}, \mathbf{s}_{\tau-k:\tau}, \mathcal{X}_{\tau-k:\tau}, \mathbf{c}_t),
\end{equation}

\begin{equation}
    \tilde{\mathbf{A}}_{\tau:\tau+W-1}
    =
    \hat{\mathbf{A}}_{\tau:\tau+W-1}
    +
    \Delta \mathbf{A}_{\tau:\tau+W-1}.
\end{equation}

Here $f_{\phi}$ is the residual Transformer with parameters $\phi$, and $\Delta \mathbf{A}_{\tau:\tau+W-1}$ is the residual action window added to the nominal lookahead window.
Unlike the nominal VLA policy, TRT operates directly on high-frequency tactile histories.
Different tactile signal types are encoded with modality-specific lightweight encoders before being fused by the residual Transformer: image-like tactile maps are processed as tactile images, matrix tactile readings are encoded with convolutional layers, and low-dimensional tactile states are embedded with Fourier features and MLPs.
At deployment, the controller executes the first $C$ actions from the corrected window and then refreshes the residual prediction with newly observed tactile feedback.
The resulting feedback rate depends on the robot control frequency, sensing pipeline, and commit interval $C$.
The policy is trained as a residual controller: the nominal action is produced by the Visuo-Tactile Goal-Conditioned Policy, and the target residual is the difference between demonstrated high-frequency actions and nominal actions in the residual action subspace.

\noindent\textbf{Why residual correction.}
Residual correction constrains the refinement policy to a focused local role.
The visuo-tactile goal-conditioned policy remains responsible for semantic and geometric progress, while the tactile-conditioned refinement policy only adjusts the nominal action to correct local contact errors.
This separation is important for fast control because tactile histories are short-range, local, and most informative near contact transitions such as slip, impact, or insertion misalignment.

\noindent\textbf{Scheduling design.}
We target a slow-to-fast hierarchy:

\begin{itemize}
    \item High-Level Planning Layer: slow semantic updates and tactile subgoal refreshes.
    \item Visuo-Tactile Goal-Conditioned Policy: intermediate nominal action-chunk generation.
    \item Tactile-Conditioned Refinement Policy: repeated residual refreshes inside the robot control loop.
\end{itemize}

These schedules are implementation choices rather than architectural constraints.
The data loader and deployment wrapper align visual, tactile, and proprioceptive histories according to the actual hardware control rate.

\section{Training}
\label{sec:training}

TouchWorld is trained in four stages.
We do not back-propagate through the entire hierarchy end to end; instead, each module is first trained with the supervision signal that matches its time scale, and the final refinement stage couples the trained nominal VLA policy with the reactive tactile layer.
This staged recipe keeps high-level reasoning, predictive goal generation, nominal action learning, and high-frequency tactile correction separable during optimization.

\subsection{Stage 1: Subtask Planner Training}

We instantiate the Subtask Planner from Qwen3-VL-4B-Instruct~\cite{yang2025qwen3technicalreport} and adapt it from annotated teleoperation traces.
Each supervision example contains the global task instruction, the current camera observation, a compact memory of recent Subtask Planner states, and the executable subtask label for the current phase.
The Subtask Planner data are sampled at the semantic update rate used for deployment and include memory variants that expose the model to previous subtasks, stale histories, and phase transitions.
The Subtask Planner dataset contains 128,866 supervision records.
The training target is the current executable subtask rather than low-level robot commands, so the fine-tuned planner learns a policy-compatible semantic interface for downstream control.

\subsection{Stage 2: Tactile World Model Training}

We instantiate the Tactile World Model from Wan2.2-TI2V-5B~\cite{wan2025wanopenadvancedlargescale} and train it separately as a predictive subgoal generator.
We first pretrain it on large-scale EgoTouch human interaction video data~\cite{zhou2026touchanythingdatasetframeworkbimanual}, which provides synchronized egocentric and wrist-mounted videos, bimanual hand pose, and dense bilateral palm pressure maps.
This pretraining stage exposes the model to diverse human-object contact patterns and teaches a general visual-to-tactile dynamics prior before robot-specific policy training.
Because the human tactile layout differs from the robot tactile glove, we use the image-form tactile representation as the shared prediction interface: EgoTouch pressure maps and robot tactile observations are both converted into visual-tactile goal grids.

After human tactile pretraining, we fine-tune the Tactile World Model on 10 hours of robot demonstrations, corresponding to approximately 1.08 million robot frames at 30 FPS.
Given the current observation grid and the selected executable subtask, the model predicts a short future visual-tactile clip or terminal subgoal state that describes the expected contact outcome.
The robot training samples are constructed from subtask-aligned demonstration segments: each sample pairs a source observation with a future 17-frame visual-tactile target at $384\times224$ resolution.
This two-stage training recipe lets the model inherit broad contact priors from human bimanual touch while adapting the final tactile subgoals to the robot embodiment and task interface.

\subsection{Stage 3: Visuo-Tactile Goal-Conditioned Policy Training}

The Visuo-Tactile Goal-Conditioned Policy is then trained as a standalone nominal action policy.
Each sample contains the global task instruction, the current executable subtask, RGB observations, image-form tactile observations, proprioceptive state, and the demonstrated future action chunk.
The policy prompt always contains both task-level context and the current subtask, as described in \cref{sec:goal_conditioned_policy}.
We train the action decoder with imitation learning and flow matching, so the policy learns to generate a nominal action chunk that advances the current subtask under visual, tactile-image, language, and proprioceptive conditioning.
When predicted visual-tactile subgoals are available, they are used as additional goal context; otherwise the policy is trained and evaluated with the current observation and subtask prompt alone.

\subsection{Stage 4: Integrated VLA-Refinement Training}

Finally, we attach the Tactile-Conditioned Refinement Policy, implemented as the reactive tactile refinement layer, to the trained VLA policy and train the coupled VLA-refinement stack on high-frequency residual targets.
For each demonstration window, the VLA policy provides a nominal action chunk $\hat{\mathbf{A}}_{t:t+H-1}$ and its action-generation context token $\mathbf{c}_t$.
For each refinement step $\tau$ inside that chunk, TRT receives the sliding nominal-action lookahead window, recent tactile history, current proprioception, and the VLA context, then predicts a residual action window $\Delta\mathbf{A}_{\tau:\tau+W-1}$.
The target residual is the difference between the demonstrated high-frequency action window and the corresponding nominal VLA lookahead window:

\begin{equation}
    \mathcal{L}_{\mathrm{fb}} =
    \left\|
    f_{\phi}(\hat{\mathbf{A}}_{\tau:\tau+W-1}, \mathbf{s}_{\tau-k:\tau}, \mathcal{X}_{\tau-k:\tau}, \mathbf{c}_t)
    -
    (\mathbf{A}^{*}_{\tau:\tau+W-1} - \hat{\mathbf{A}}_{\tau:\tau+W-1})
    \right\|_2^2.
\end{equation}

Here $\mathcal{L}_{\mathrm{fb}}$ is the feedback-refinement loss and $\mathbf{A}^{*}_{\tau:\tau+W-1}$ is the demonstrated high-frequency action window.
The loss is applied in the tactile-sensitive residual action subspace, while action dimensions outside this subspace follow the nominal VLA output.
This final stage trains the refinement layer in the same action context used at deployment while preserving the nominal policy's responsibility for semantic and geometric progress.
\begin{figure}[t]
\centering
\includegraphics[width=\linewidth]{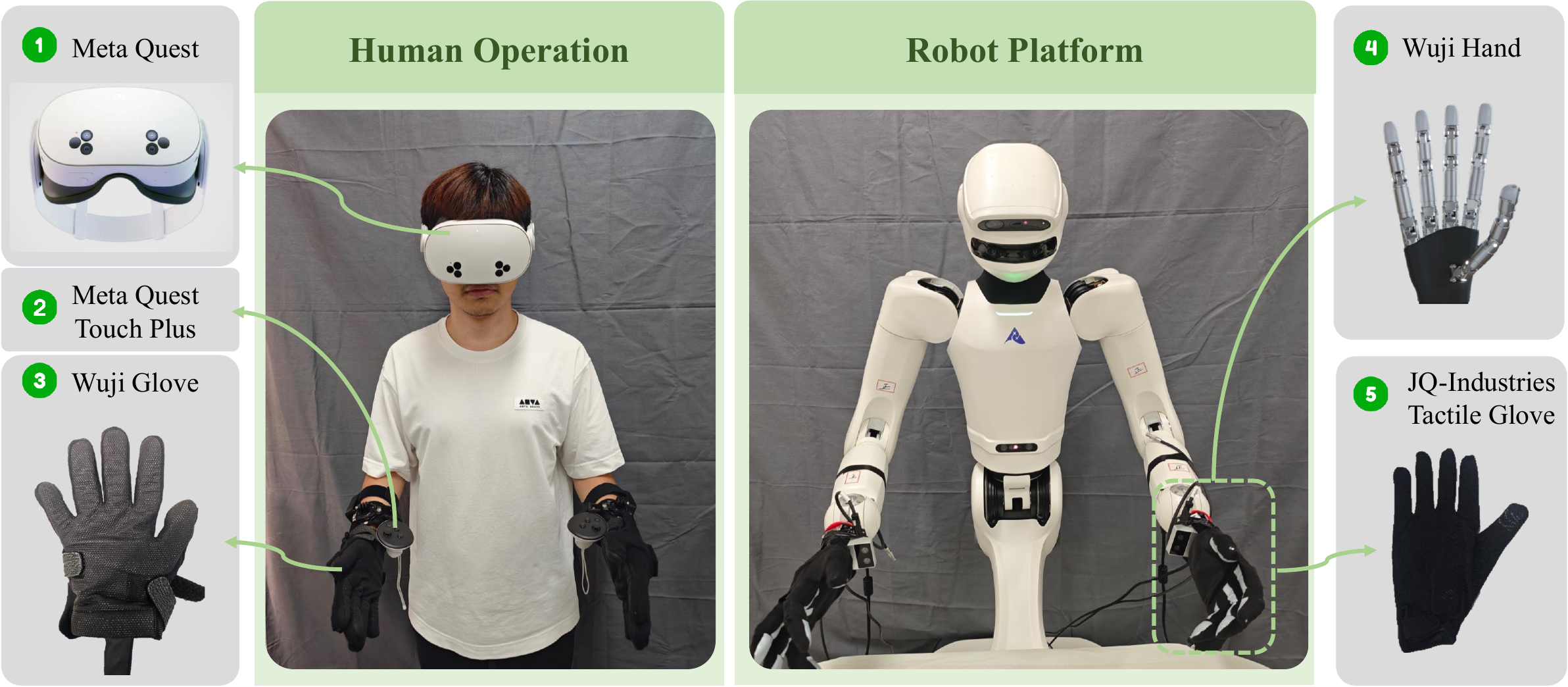}
\caption{Hardware platform for TouchWorld. The human operator provides visual and hand-motion inputs through a Meta Quest headset, Meta Quest Touch Plus controllers, and a Wuji Glove. The robot platform uses Wuji dexterous hands and a JQ-Industries tactile glove for contact-rich manipulation with tactile feedback.}
\label{fig:hardware}
\end{figure}
\section{Experiments}
\label{sec:experiments}

We evaluate TouchWorld on real-robot manipulation tasks that require long-horizon reasoning, contact-aware action generation, and recovery from local contact disturbances.
The experiments answer four questions:
(1) whether TouchWorld improves task success over strong generalist and tactile-policy baselines;
(2) which predictive and reactive components contribute most to robustness;
(3) whether the Tactile World Model predicts tactile subgoals that match real contact outcomes; and
(4) whether the Subtask Planner produces subtasks that are executable by the downstream policy.

\subsection{Experimental Setup}

The hardware setup is shown in \cref{fig:hardware}.
On the teleoperation side, the demonstrator wears a Meta Quest headset for immersive visual feedback and uses Meta Quest Touch Plus controllers together with a Wuji Glove to provide hand-motion commands.
On the robot side, a humanoid platform equipped with Wuji dexterous hands executes the manipulation policy, while a JQ-Industries tactile glove mounted on the hand provides contact observations for tactile-conditioned action generation and refinement.
This setup supports teleoperated data collection and policy evaluation under the same visual, proprioceptive, and tactile sensing interfaces.

\noindent\textbf{Policy scheduling.}
In our implementation, the nominal VLA policy predicts action chunks with $H=32$.
The refinement layer uses a sliding nominal-action lookahead window with $W=16$ and commits the first $C=4$ corrected actions before refreshing the residual prediction.
For example, relative to a nominal chunk, before executing step 0, TRT conditions on nominal actions $[0,15]$ and commits corrected actions $[0,3]$; before step 4, it conditions on nominal actions $[4,19]$ and commits corrected actions $[4,7]$.
These values are implementation hyperparameters rather than architectural constraints, and the resulting feedback rate depends on the robot control frequency and sensing pipeline.

\begin{figure}[t]
\centering
\includegraphics[width=\linewidth]{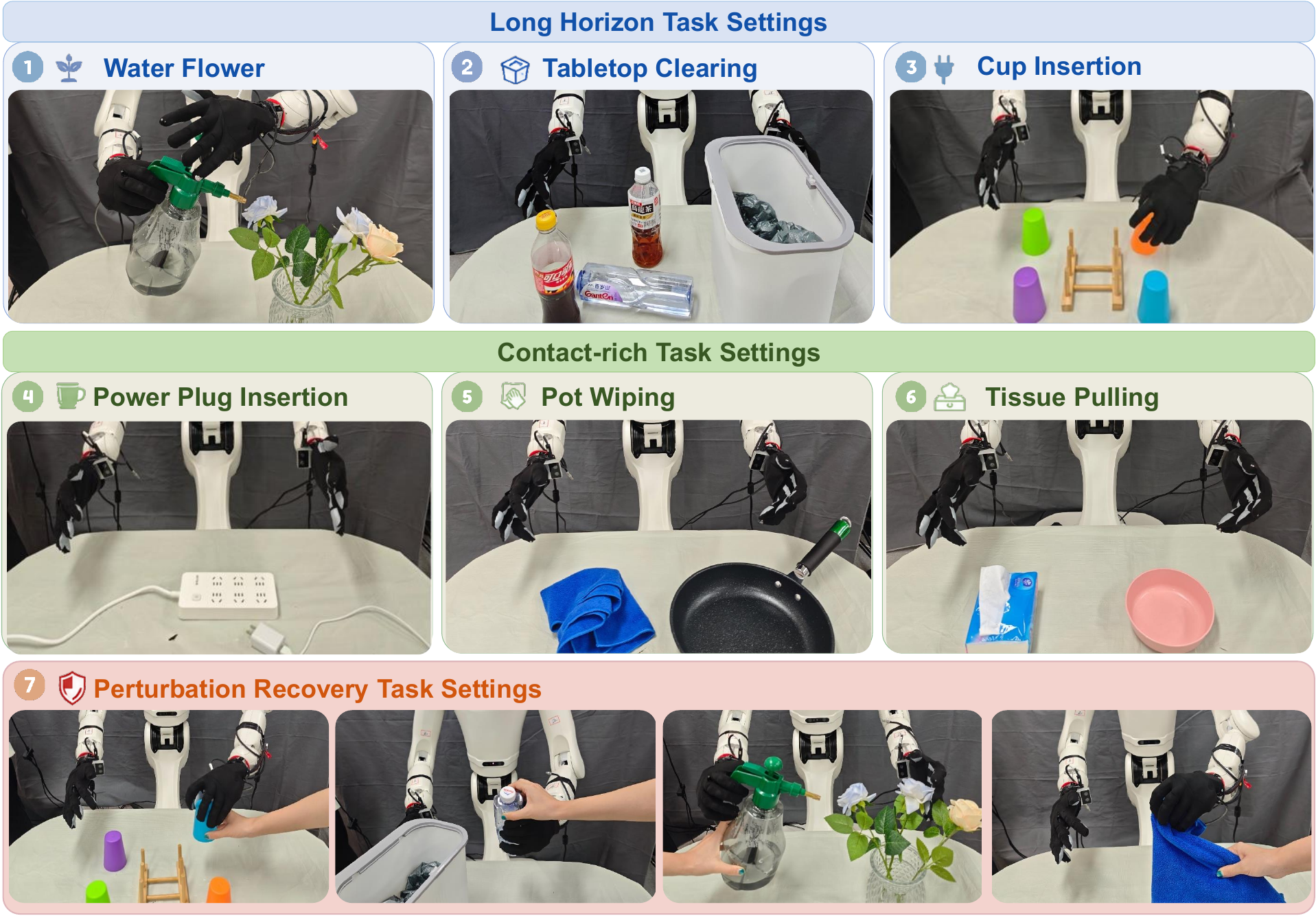}
\caption{Real-robot task suite for evaluating TouchWorld. The six tasks cover long-horizon planning, precision insertion, continuous contact regulation, soft-object handling, and recovery from human perturbations.}
\label{fig:task_suite}
\end{figure}

We evaluate on a six-task real-robot suite that spans different failure modes of tactile manipulation.
As shown in \cref{fig:task_suite}, the six tasks include long-horizon semantic manipulation, such as \textit{Water Flower} and \textit{Tabletop Clearing}; precision contact tasks, such as \textit{Cup Insertion} and \textit{Power Plug Insertion}; continuous contact regulation in \textit{Pot Wiping}; and soft-object handling in \textit{Tissue Pulling}.
Each task is evaluated in a clean setting and a human perturbation setting.
In the perturbation setting, the operator applies disturbances such as target displacement, unstable contact, or grasp interference during execution.
This setup tests the intended roles of TouchWorld: semantic decomposition by the Subtask Planner, predictive contact goals from the Tactile World Model, nominal action generation by the Visuo-Tactile Goal-Conditioned Policy, and high-frequency correction by the Tactile-Conditioned Refinement Policy.
For each task, we collect 200 teleoperated training trajectories and conduct 100 real-robot evaluation rollouts.

For real-robot experiments, we report success rate over repeated rollouts in both settings.
The Wuji data pipeline exports teleoperated trajectories for the evaluated tasks.
Each trajectory is exported with global task instructions, manually annotated subtask segments, head and wrist camera streams, proprioceptive states, actions, and left/right tactile pressure observations.
These annotations support subtask-conditioned policy training, Tactile World Model supervision, residual refinement training, and Subtask Planner evaluation.

\subsection{Baselines}

For the main real-robot comparison, we evaluate TouchWorld against three policy baselines:

\begin{itemize}
    \item \textbf{Pi-0.5}: a vision-language-action policy baseline evaluated with the same task instructions and visual observations.
    \item \textbf{FTP-1}: the previous monolithic tactile policy baseline without the proposed predictive-and-reactive hierarchy.
    \item \textbf{GR00T N1.7}: a generalist robot policy baseline evaluated on the same real-robot task suite.
\end{itemize}

Component ablations are reported separately in \cref{fig:ablations}.

\subsection{Main Results}
\begin{table*}[t]
\centering
\caption{Per-task manipulation success rates (\%) on the real-robot TouchWorld benchmark. Each task is evaluated in both a clean setting and a human perturbation setting. The best and second-best results in each setting are highlighted in \textbf{bold} and \underline{underline}, respectively.}
\vspace{-1mm}
\label{tab:main_results}
\setlength{\tabcolsep}{4.2pt}
\renewcommand{\arraystretch}{1.08}
\resizebox{\textwidth}{!}{%
\begin{tabular}{@{}lccccccc@{}}
\toprule
\textbf{Method}
& \textbf{Water Flower}
& \textbf{Tabletop Clearing}
& \textbf{Cup Insertion}
& \textbf{Power Plug Insertion}
& \textbf{Pot Wiping}
& \textbf{Tissue Pulling}
& \textbf{Avg.} \\
\midrule
\rowcolor{categorygreen}
\multicolumn{8}{l}{\normalsize\textit{Clean Setting}} \\
\textbf{Pi-0.5}
& 52 & \underline{66} & 36 & 12 & 39 & 39 & 40.7 \\
\textbf{FTP-1}
& \underline{56} & 60 & \underline{48} & \underline{32} & \underline{57} & \underline{43} & \underline{49.3} \\
\textbf{GR00T N1.7}
& 50 & 58 & 33 & 18 & 36 & 41 & 39.3 \\
\midrule
\rowcolor{oursblue}
\textbf{TouchWorld}
& \textbf{72} & \textbf{76} & \textbf{66} & \textbf{45} & \textbf{70} & \textbf{61} & \textbf{65.0} \\
\midrule
\rowcolor{categoryorange}
\multicolumn{8}{l}{\normalsize\textit{Human Perturbation Setting}} \\
\textbf{Pi-0.5}
& 34 & \underline{44} & 24 & 6 & 28 & 30 & 27.7 \\
\textbf{FTP-1}
& \underline{39} & 42 & \underline{34} & \underline{20} & \underline{42} & \underline{34} & \underline{35.2} \\
\textbf{GR00T N1.7}
& 32 & 36 & 21 & 9 & 26 & 32 & 26.0 \\
\midrule
\rowcolor{oursblue}
\textbf{TouchWorld}
& \textbf{60} & \textbf{62} & \textbf{52} & \textbf{35} & \textbf{57} & \textbf{56} & \textbf{53.7} \\
\bottomrule
\end{tabular}
}
\end{table*}

The results in \cref{tab:main_results} show that TouchWorld consistently improves over the baselines across both clean and human-perturbed rollouts.
In the clean setting, TouchWorld reaches 65.0\% average success, improving over the strongest baseline by 15.7 percentage points.
Under human perturbations, TouchWorld reaches 53.7\% average success, improving over the strongest baseline by 18.5 percentage points.
The gains are especially clear on power plug insertion, pot wiping, and tissue pulling, where tactile prediction and fast local correction are most important.
TouchWorld also improves water flower and tabletop clearing, indicating that the hierarchy preserves high-level task execution while strengthening contact-aware manipulation.

\subsection{Ablation Studies}

\begin{figure}[t]
\centering
\includegraphics[width=\linewidth]{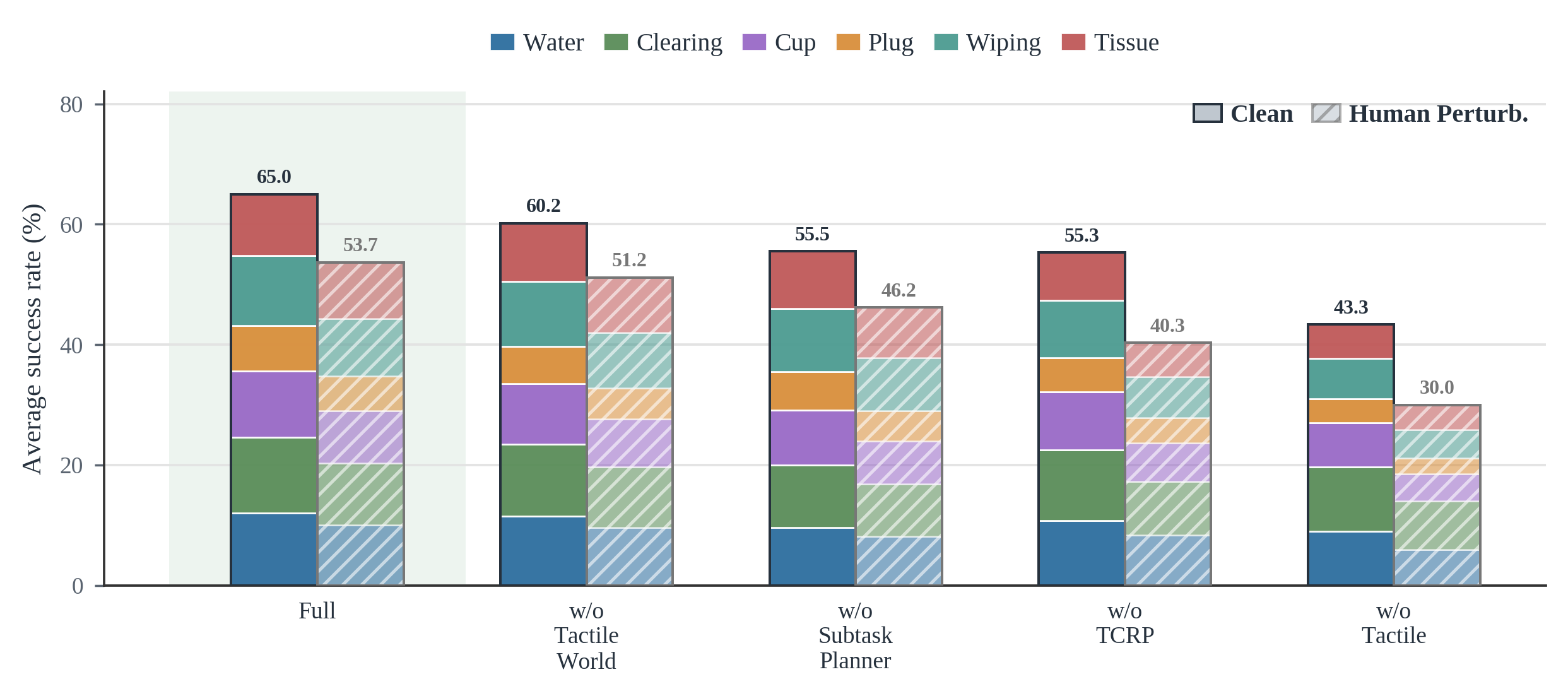}
\caption{Stacked ablation results for TouchWorld. Each bar reports the six-task average success rate, with colored segments showing the contribution of each task to the average. Solid bars denote clean rollouts, while hatched bars denote human-perturbed rollouts that require corrective behavior.}
\label{fig:ablations}
\end{figure}

\Cref{fig:ablations} summarizes how the main interface and feedback choices affect the average performance across the six-task suite.
The stacked visualization shows both the overall success rate and the task composition behind each average, making it easier to inspect whether a component mainly affects long-horizon, precision-contact, or soft-object tasks.
We ablate the following design choices:

\begin{itemize}
    \item \textbf{Subtask Planner}: remove subtask decomposition and condition the policy directly on the original task prompt.
    \item \textbf{Tactile World Model}: remove future visual-tactile goal prediction and condition the policy only on the current observation and current subtask.
    \item \textbf{Tactile-Conditioned Refinement Policy}: remove the residual feedback layer and execute nominal action chunks directly.
    \item \textbf{Tactile input}: remove tactile observations from the policy input to measure the contribution of touch sensing.
\end{itemize}

Removing tactile input causes the largest degradation, confirming that contact observations are essential for this benchmark.
Removing the Tactile-Conditioned Refinement Policy particularly hurts the human perturbation setting, where the system must correct local execution errors online.
Removing the Subtask Planner mainly reduces long-horizon consistency, while removing the Tactile World Model weakens contact-aware goal conditioning.

\subsection{Tactile World Model Prediction Analysis}

We further evaluate whether the Tactile World Model predicts tactile subgoals that match the contact state reached by the robot.
For each held-out trajectory, we sample a source observation and executable subtask, generate a 17-frame visual-tactile goal clip, and align it with the real future segment in the same rollout.
The ground-truth tactile goal is the tactile pressure observation captured when the robot reaches the annotated subgoal state for that subtask.
We evaluate the predicted tactile sequence with temporal contact accuracy over the 17-frame window, and evaluate the terminal tactile goal with contact IoU and volumetric IoU after thresholding the pressure map at $\tau$.
We compare against a persistence baseline that copies the current tactile observation as the future goal and a nearest-neighbor baseline that retrieves a subgoal from the training set using the task and subtask label.

\begin{table}[t]
\centering
\caption{Tactile World Model prediction accuracy on held-out subgoal segments. Ground truth is the tactile pressure observation collected when the robot reaches the corresponding annotated subgoal state. All numbers are percentages, and higher is better for all metrics.}
\vspace{-1mm}
\label{tab:twm_prediction}
\setlength{\tabcolsep}{5pt}
\renewcommand{\arraystretch}{1.12}
\small
\begin{tabular}{@{}lccc@{}}
\toprule
\textbf{Method}
& \textbf{\shortstack{Temporal\\Contact Acc.}}
& \textbf{\shortstack{Contact\\IoU}}
& \textbf{\shortstack{Volumetric\\IoU}} \\
\midrule
Current tactile copy
& 70.4 & 31.8 & 24.6 \\
Nearest-neighbor subgoal
& \underline{77.5} & \underline{39.2} & \underline{31.0} \\
\rowcolor{oursblue}
\textbf{Tactile World Model}
& \textbf{86.3} & \textbf{52.7} & \textbf{43.8} \\
\bottomrule
\end{tabular}
\end{table}

\Cref{tab:twm_prediction} shows that the Tactile World Model predicts substantially more accurate contact timing and terminal tactile geometry than simple persistence or retrieval baselines.
This supports its role as a predictive contact reference for the downstream goal-conditioned policy.

\subsection{Vision-Language Subtask Planner Analysis}

We analyze whether the Vision-Language Subtask Planner produces subtasks that are both semantically correct and executable by the downstream policy.
A Subtask Planner output can fail in two ways:
(1) it may choose an incorrect subtask for the scene, or
(2) it may choose a correct subtask that lies outside the policy's trained behavior distribution.
We therefore report subtask correctness, downstream execution success, and transition F1 over task phases.
\begin{table}[t]
\centering
\caption{Evaluation of Vision-Language Subtask Planner variants. We compare zero-shot planners, supervised fine-tuning without memory, and the memory-augmented planner used by TouchWorld. Metrics are reported in percentage.}
\vspace{-1mm}
\label{tab:planner_metrics}
\setlength{\tabcolsep}{5pt}
\renewcommand{\arraystretch}{1.12}
\small
\begin{tabular}{@{}lccc@{}}
\toprule
\textbf{Planner}
& \textbf{\shortstack{Subtask\\Acc.}}
& \textbf{\shortstack{Execution\\Success}}
& \textbf{\shortstack{Transition\\F1}} \\
\midrule
Zero-shot Qwen3-VL-4B~\cite{yang2025qwen3technicalreport}
& 43 & 34 & 62 \\

Zero-shot Qwen3-VL-32B~\cite{yang2025qwen3technicalreport}
& 69 & 54 & 71 \\

SFT Qwen3-VL-4B w/o Memory
& \underline{73} & \underline{60} & \underline{76} \\

\rowcolor{oursblue}
\textbf{Memory-Augmented SFT Qwen3-VL-4B (Ours)}
& \textbf{88} & \textbf{65} & \textbf{82} \\
\bottomrule
\end{tabular}
\end{table}

\Cref{tab:planner_metrics} shows that supervised subtask tuning improves over zero-shot planning, and that Subtask Planner memory further improves phase-transition consistency.
The memory-augmented 4B Subtask Planner outperforms the zero-shot 32B planner, suggesting that task-phase supervision and execution history are more important than model scale alone for this interface.

%

\subsection{Qualitative Inference Analysis}

\begin{figure}[t]
\centering
\includegraphics[width=\linewidth]{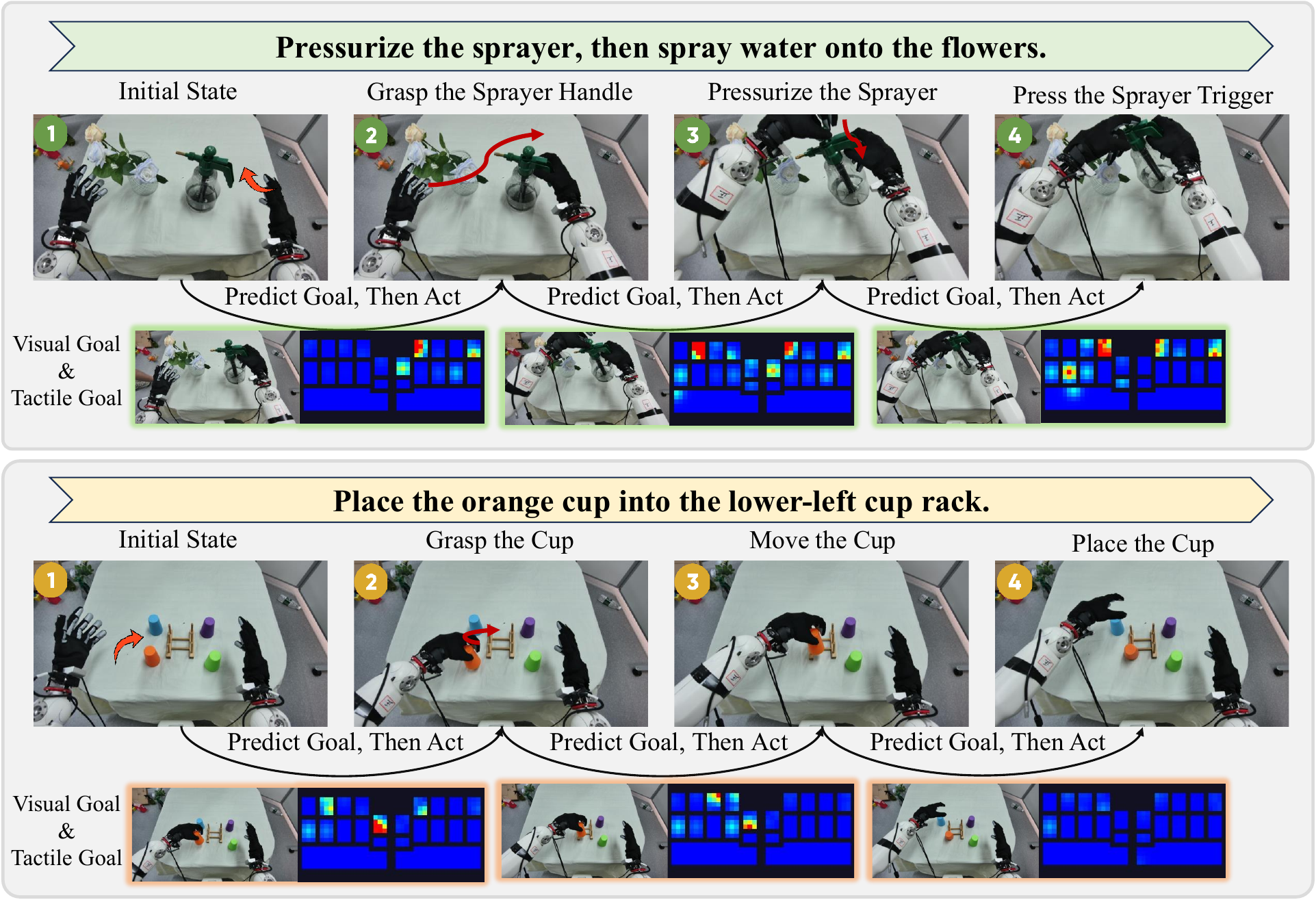}
\caption{Qualitative inference demonstration of TouchWorld. For each task, the model progresses from the initial scene through a sequence of executable subtasks and predicts tactile subgoals that support contact-aware manipulation.}
\label{fig:inference_demo}
\end{figure}

We further visualize the inference process on representative long-horizon manipulation tasks.
As shown in \cref{fig:inference_demo}, TouchWorld decomposes each instruction into executable intermediate subtasks and predicts tactile subgoals that provide contact-aware references for downstream action generation.

\section{Discussion and Limitations}

TouchWorld demonstrates that tactile prediction and tactile feedback refinement can be combined in a practical manipulation policy, but several limitations remain.
First, our real-robot evaluation focuses on six representative contact-rich tasks.
These tasks cover planning, insertion, wiping, and soft-object handling, but they do not yet exhaust the diversity of household manipulation or deformable-object interactions.
Scaling the benchmark to more objects, workspaces, and task compositions is an important next step.

Second, the Tactile World Model currently predicts short-horizon visual-tactile subgoals.
This is sufficient for contact-aware subtask execution, but longer-horizon prediction remains challenging when object motion, hand occlusion, or human perturbations produce multiple plausible futures.
Future work can improve this component with uncertainty-aware subgoal prediction or by generating multiple candidate tactile subgoals for downstream selection.

Third, TouchWorld is tied to the sensing layout used in our robot platform.
The policy uses image-form tactile observations in the nominal VLA branch and high-frequency tactile histories in the refinement branch.
Although this design is modular, transferring to a different tactile sensor or hand morphology still requires calibration, normalization, and likely a small amount of adaptation data.

Finally, the hierarchy introduces scheduling hyperparameters, including the High-Level Planning Layer update rate, the Tactile World Model refresh rule, the nominal action chunk length, and the residual commit interval.
We use a fixed scheduling profile in this work, which worked well for our tasks.
More adaptive scheduling could further reduce computation and improve responsiveness when contact dynamics change rapidly.

\section{Conclusion}

We presented \textbf{TouchWorld}, a predictive-and-reactive tactile foundation model for dexterous manipulation.
TouchWorld is implemented as a hierarchical manipulation policy that combines a High-Level Planning Layer, which contains vision-language subtask planning and Tactile World Model prediction, with visuo-tactile goal-conditioned action generation and high-frequency tactile residual refinement.
This design treats touch as both a predictive cue for contact-aware goal generation and a fast feedback signal for online correction.
Across six real-robot tasks, TouchWorld improves average success in both clean and human-perturbed settings, showing the benefit of explicitly separating semantic planning, predictive tactile subgoals, nominal action generation, and reactive tactile feedback.
Future work will scale the task suite, improve uncertainty-aware tactile goal prediction, and study broader transfer across tactile sensors and robot hands.



\clearpage

\bibliographystyle{plainnat}
\bibliography{references}

\begin{thebibliography}{35}
\providecommand{\natexlab}[1]{#1}
\providecommand{\url}[1]{\texttt{#1}}
\expandafter\ifx\csname urlstyle\endcsname\relax
  \providecommand{\doi}[1]{doi: #1}\else
  \providecommand{\doi}{doi: \begingroup \urlstyle{rm}\Url}\fi

\bibitem[Black et~al.(2025)Black, Galliker, and
  Levine]{black2025realtimeexecutionactionchunking}
Kevin Black, Manuel~Y. Galliker, and Sergey Levine.
\newblock Real-time execution of action chunking flow policies, 2025.
\newblock URL \url{https://arxiv.org/abs/2506.07339}.

\bibitem[Black et~al.(2026)Black, Brown, Driess, Esmail, Equi, Finn, Fusai,
  Groom, Hausman, Ichter, Jakubczak, Jones, Ke, Levine, Li-Bell, Mothukuri,
  Nair, Pertsch, Shi, Tanner, Vuong, Walling, Wang, and
  Zhilinsky]{black2026pi0visionlanguageactionflowmodel}
Kevin Black, Noah Brown, Danny Driess, Adnan Esmail, Michael Equi, Chelsea
  Finn, Niccolo Fusai, Lachy Groom, Karol Hausman, Brian Ichter, Szymon
  Jakubczak, Tim Jones, Liyiming Ke, Sergey Levine, Adrian Li-Bell, Mohith
  Mothukuri, Suraj Nair, Karl Pertsch, Lucy~Xiaoyang Shi, James Tanner, Quan
  Vuong, Anna Walling, Haohuan Wang, and Ury Zhilinsky.
\newblock $\pi_0$: A vision-language-action flow model for general robot
  control, 2026.
\newblock URL \url{https://arxiv.org/abs/2410.24164}.

\bibitem[Chi et~al.(2025)Chi, Xu, Feng, Cousineau, Du, Burchfiel, Tedrake, and
  Song]{chi2025diffusion}
Cheng Chi, Zhenjia Xu, Siyuan Feng, Eric Cousineau, Yilun Du, Benjamin
  Burchfiel, Russ Tedrake, and Shuran Song.
\newblock Diffusion policy: Visuomotor policy learning via action diffusion.
\newblock \emph{The International Journal of Robotics Research}, 44\penalty0
  (10-11):\penalty0 1684--1704, 2025.

\bibitem[Dai et~al.(2026)Dai, Lan, Zhou, Zhao, Su, Tong, Guan, and
  Yang]{dai2026conlacontrastivelatentaction}
Weisheng Dai, Kai Lan, Jianyi Zhou, Bo~Zhao, Xiu Su, Junwen Tong, Weili Guan,
  and Shuo Yang.
\newblock Conla: Contrastive latent action learning from human videos for
  robotic manipulation, 2026.
\newblock URL \url{https://arxiv.org/abs/2602.00557}.

\bibitem[Grady et~al.(2024)]{grady2024pressurevision++}
Patrick Grady et~al.
\newblock Pressurevision++: Estimating fingertip pressure from diverse rgb
  images.
\newblock In \emph{IEEE/CVF Conference on Computer Vision and Pattern
  Recognition (CVPR)}, 2024.

\bibitem[Huang et~al.(2025)Huang, Wang, Lin, Hu, Wen, and
  Gao]{huang2025tactilevlaunlockingvisionlanguageactionmodels}
Jialei Huang, Shuo Wang, Fanqi Lin, Yihang Hu, Chuan Wen, and Yang Gao.
\newblock Tactile-vla: Unlocking vision-language-action model's physical
  knowledge for tactile generalization, 2025.
\newblock URL \url{https://arxiv.org/abs/2507.09160}.

\bibitem[Huang et~al.(2026)Huang, Ye, Gong, Zhu, Gao, and
  Zhang]{huang2026spatiallyanchoredtactileawareness}
Jialei Huang, Yang Ye, Yuanqing Gong, Xuezhou Zhu, Yang Gao, and Kaifeng Zhang.
\newblock Spatially anchored tactile awareness for robust dexterous
  manipulation, 2026.
\newblock URL \url{https://arxiv.org/abs/2510.14647}.

\bibitem[Intelligence et~al.(2025{\natexlab{a}})Intelligence, Amin, Aniceto,
  Balakrishna, Black, Conley, Connors, Darpinian, Dhabalia, DiCarlo, Driess,
  Equi, Esmail, Fang, Finn, Glossop, Godden, Goryachev, Groom, Hancock,
  Hausman, Hussein, Ichter, Jakubczak, Jen, Jones, Katz, Ke, Kuchi, Lamb,
  LeBlanc, Levine, Li-Bell, Lu, Mano, Mothukuri, Nair, Pertsch, Ren, Sharma,
  Shi, Smith, Springenberg, Stachowicz, Stoeckle, Swerdlow, Tanner, Torne,
  Vuong, Walling, Wang, Williams, Yoo, Yu, Zhilinsky, and
  Zhou]{intelligence2025pi06vlalearnsexperience}
Physical Intelligence, Ali Amin, Raichelle Aniceto, Ashwin Balakrishna, Kevin
  Black, Ken Conley, Grace Connors, James Darpinian, Karan Dhabalia, Jared
  DiCarlo, Danny Driess, Michael Equi, Adnan Esmail, Yunhao Fang, Chelsea Finn,
  Catherine Glossop, Thomas Godden, Ivan Goryachev, Lachy Groom, Hunter
  Hancock, Karol Hausman, Gashon Hussein, Brian Ichter, Szymon Jakubczak, Rowan
  Jen, Tim Jones, Ben Katz, Liyiming Ke, Chandra Kuchi, Marinda Lamb, Devin
  LeBlanc, Sergey Levine, Adrian Li-Bell, Yao Lu, Vishnu Mano, Mohith
  Mothukuri, Suraj Nair, Karl Pertsch, Allen~Z. Ren, Charvi Sharma,
  Lucy~Xiaoyang Shi, Laura Smith, Jost~Tobias Springenberg, Kyle Stachowicz,
  Will Stoeckle, Alex Swerdlow, James Tanner, Marcel Torne, Quan Vuong, Anna
  Walling, Haohuan Wang, Blake Williams, Sukwon Yoo, Lili Yu, Ury Zhilinsky,
  and Zhiyuan Zhou.
\newblock $\pi^{*}_{0.6}$: a vla that learns from experience,
  2025{\natexlab{a}}.
\newblock URL \url{https://arxiv.org/abs/2511.14759}.

\bibitem[Intelligence et~al.(2025{\natexlab{b}})Intelligence, Black, Brown,
  Darpinian, Dhabalia, Driess, Esmail, Equi, Finn, Fusai, Galliker, Ghosh,
  Groom, Hausman, Ichter, Jakubczak, Jones, Ke, LeBlanc, Levine, Li-Bell,
  Mothukuri, Nair, Pertsch, Ren, Shi, Smith, Springenberg, Stachowicz, Tanner,
  Vuong, Walke, Walling, Wang, Yu, and
  Zhilinsky]{intelligence2025pi05visionlanguageactionmodelopenworld}
Physical Intelligence, Kevin Black, Noah Brown, James Darpinian, Karan
  Dhabalia, Danny Driess, Adnan Esmail, Michael Equi, Chelsea Finn, Niccolo
  Fusai, Manuel~Y. Galliker, Dibya Ghosh, Lachy Groom, Karol Hausman, Brian
  Ichter, Szymon Jakubczak, Tim Jones, Liyiming Ke, Devin LeBlanc, Sergey
  Levine, Adrian Li-Bell, Mohith Mothukuri, Suraj Nair, Karl Pertsch, Allen~Z.
  Ren, Lucy~Xiaoyang Shi, Laura Smith, Jost~Tobias Springenberg, Kyle
  Stachowicz, James Tanner, Quan Vuong, Homer Walke, Anna Walling, Haohuan
  Wang, Lili Yu, and Ury Zhilinsky.
\newblock $\pi_{0.5}$: a vision-language-action model with open-world
  generalization, 2025{\natexlab{b}}.
\newblock URL \url{https://arxiv.org/abs/2504.16054}.

\bibitem[Intelligence et~al.(2026)Intelligence, Ai, Amin, Aniceto, Balakrishna,
  Balke, Black, Bokinsky, Cao, Charbonnier, Choudhary, Collins, Conley,
  Connors, Darpinian, Dhabalia, Dhaka, DiCarlo, Driess, Equi, Esmail, Fang,
  Finn, Glossop, Godden, Goryachev, Groom, Habeeb, Hancock, Hausman, Hussein,
  Hwang, Ichter, Jacobsen, Jakubczak, Jen, Jones, Kammerer, Katz, Ke, Khadikov,
  Kuchi, Lamb, LeBlanc, LeCount, Levine, Li, Li-Bell, Lialin, Liang, Lim, Lu,
  Luo, Mano, Marwaha, Mongush, Murphy, Nair, Patterson, Pertsch, Ren, Schelske,
  Sharma, Shi, Shi, Smith, Springenberg, Stachowicz, Stoeckle, Tang, Tanner,
  Tekeste, Torne, Vedder, Vuong, Walling, Wang, Wang, Wang, Whalen, Whitmore,
  Williams, Xu, Yoo, Yu, Zhang, Zhang, and
  Zhilinsky]{intelligence2026pi07steerablegeneralistrobotic}
Physical Intelligence, Bo~Ai, Ali Amin, Raichelle Aniceto, Ashwin Balakrishna,
  Greg Balke, Kevin Black, George Bokinsky, Shihao Cao, Thomas Charbonnier,
  Vedant Choudhary, Foster Collins, Ken Conley, Grace Connors, James Darpinian,
  Karan Dhabalia, Maitrayee Dhaka, Jared DiCarlo, Danny Driess, Michael Equi,
  Adnan Esmail, Yunhao Fang, Chelsea Finn, Catherine Glossop, Thomas Godden,
  Ivan Goryachev, Lachlan Groom, Haroun Habeeb, Hunter Hancock, Karol Hausman,
  Gashon Hussein, Victor Hwang, Brian Ichter, Connor Jacobsen, Szymon
  Jakubczak, Rowan Jen, Tim Jones, Gregg Kammerer, Ben Katz, Liyiming Ke,
  Mairbek Khadikov, Chandra Kuchi, Marinda Lamb, Devin LeBlanc, Brendon
  LeCount, Sergey Levine, Xinyu Li, Adrian Li-Bell, Vladislav Lialin, Zhonglin
  Liang, Wallace Lim, Yao Lu, Enyu Luo, Vishnu Mano, Nandan Marwaha, Aikys
  Mongush, Liam Murphy, Suraj Nair, Tyler Patterson, Karl Pertsch, Allen~Z.
  Ren, Gavin Schelske, Charvi Sharma, Baifeng Shi, Lucy~Xiaoyang Shi, Laura
  Smith, Jost~Tobias Springenberg, Kyle Stachowicz, Will Stoeckle, Jiaming
  Tang, Jimmy Tanner, Shalom Tekeste, Marcel Torne, Kyle Vedder, Quan Vuong,
  Anna Walling, Haohuan Wang, Jason Wang, XuDong Wang, Chris Whalen, Samuel
  Whitmore, Blake Williams, Charles Xu, Sukwon Yoo, Lili Yu, Wuming Zhang,
  Zhuoyang Zhang, and Ury Zhilinsky.
\newblock ${\pi}_{0.7}$: a steerable generalist robotic foundation model with
  emergent capabilities, 2026.
\newblock URL \url{https://arxiv.org/abs/2604.15483}.

\bibitem[Kim et~al.(2024)Kim, Pertsch, Karamcheti, Xiao, Balakrishna, Nair,
  Rafailov, Foster, Lam, Sanketi, et~al.]{kim2024openvla}
Moo~Jin Kim, Karl Pertsch, Siddharth Karamcheti, Ted Xiao, Ashwin Balakrishna,
  Suraj Nair, Rafael Rafailov, Ethan Foster, Grace Lam, Pannag Sanketi, et~al.
\newblock Openvla: An open-source vision-language-action model.
\newblock \emph{arXiv preprint arXiv:2406.09246}, 2024.

\bibitem[Kim et~al.(2025)Kim, Finn, and
  Liang]{kim2025finetuningvisionlanguageactionmodelsoptimizing}
Moo~Jin Kim, Chelsea Finn, and Percy Liang.
\newblock Fine-tuning vision-language-action models: Optimizing speed and
  success, 2025.
\newblock URL \url{https://arxiv.org/abs/2502.19645}.

\bibitem[Lambeta et~al.(2020)Lambeta, Chou, Tian, et~al.]{lambeta2020digit}
Mike Lambeta, Po-Wei Chou, Stephen Tian, et~al.
\newblock Digit: A novel design for a low-cost compact high-resolution tactile
  sensor with application to in-hand manipulation.
\newblock In \emph{IEEE Robotics and Automation Letters}, 2020.

\bibitem[Li et~al.(2025)Li, Deng, Zhang, Jang, Memmel, Yu, Garrett, Ramos, Fox,
  Li, Gupta, and Goyal]{li2025hamsterhierarchicalactionmodels}
Yi~Li, Yuquan Deng, Jesse Zhang, Joel Jang, Marius Memmel, Raymond Yu,
  Caelan~Reed Garrett, Fabio Ramos, Dieter Fox, Anqi Li, Abhishek Gupta, and
  Ankit Goyal.
\newblock Hamster: Hierarchical action models for open-world robot
  manipulation, 2025.
\newblock URL \url{https://arxiv.org/abs/2502.05485}.

\bibitem[Li et~al.(2019)Li, Li, Torralba, et~al.]{li2019visgel}
Yunzhu Li, Jun-Yan Li, Antonio Torralba, et~al.
\newblock Connecting touch and vision via cross-modal prediction.
\newblock In \emph{IEEE/CVF Conference on Computer Vision and Pattern
  Recognition (CVPR)}, 2019.

\bibitem[Liu et~al.(2025)Liu, Wu, Li, Tan, Chen, Wang, Xu, Su, and
  Zhu]{liu2025rdt1bdiffusionfoundationmodel}
Songming Liu, Lingxuan Wu, Bangguo Li, Hengkai Tan, Huayu Chen, Zhengyi Wang,
  Ke~Xu, Hang Su, and Jun Zhu.
\newblock Rdt-1b: a diffusion foundation model for bimanual manipulation, 2025.
\newblock URL \url{https://arxiv.org/abs/2410.07864}.

\bibitem[Niu et~al.(2026)Niu, Liu, Wang, Shao, Yin, Pai, Sharma, Saravalle,
  Zheng, Wang, Punamiya, Xu, Xie, Jiang, Fu, Kallidromitis, Gioia, Zhang, Ge,
  Feng, Galasso, Zhan, Chan, Bai, Herzig, Lei, Fei-Fei, Goldberg, Malik,
  Abbeel, Zhu, Xu, Fan, and
  Darrell]{niu2026trextactilereactivedexterousmanipulation}
Dantong Niu, Zhuoyang Liu, Zekai Wang, Boning Shao, Zhao-Heng Yin, Anirudh Pai,
  Yuvan Sharma, Stefano Saravalle, Ruijie Zheng, Jing Wang, Ryan Punamiya,
  Mengda Xu, Yuqi Xie, Yunfan Jiang, Letian Fu, Konstantinos Kallidromitis,
  Matteo Gioia, Junyi Zhang, Jiaxin Ge, Haiwen Feng, Fabio Galasso, Wei Zhan,
  David~M. Chan, Yutong Bai, Roei Herzig, Jiahui Lei, Li~Fei-Fei, Ken Goldberg,
  Jitendra Malik, Pieter Abbeel, Yuke Zhu, Danfei Xu, Linxi Fan, and Trevor
  Darrell.
\newblock T-rex: Tactile-reactive dexterous manipulation, 2026.
\newblock URL \url{https://arxiv.org/abs/2606.17055}.

\bibitem[NVIDIA et~al.(2025)NVIDIA, :, Bjorck, Castañeda, Cherniadev, Da,
  Ding, Fan, Fang, Fox, Hu, Huang, Jang, Jiang, Kautz, Kundalia, Lao, Li, Lin,
  Lin, Liu, Llontop, Magne, Mandlekar, Narayan, Nasiriany, Reed, Tan, Wang,
  Wang, Wang, Wang, Xiang, Xie, Xu, Xu, Ye, Yu, Zhang, Zhang, Zhao, Zheng, and
  Zhu]{nvidia2025gr00tn1openfoundation}
NVIDIA, :, Johan Bjorck, Fernando Castañeda, Nikita Cherniadev, Xingye Da,
  Runyu Ding, Linxi~"Jim" Fan, Yu~Fang, Dieter Fox, Fengyuan Hu, Spencer Huang,
  Joel Jang, Zhenyu Jiang, Jan Kautz, Kaushil Kundalia, Lawrence Lao, Zhiqi Li,
  Zongyu Lin, Kevin Lin, Guilin Liu, Edith Llontop, Loic Magne, Ajay Mandlekar,
  Avnish Narayan, Soroush Nasiriany, Scott Reed, You~Liang Tan, Guanzhi Wang,
  Zu~Wang, Jing Wang, Qi~Wang, Jiannan Xiang, Yuqi Xie, Yinzhen Xu, Zhenjia Xu,
  Seonghyeon Ye, Zhiding Yu, Ao~Zhang, Hao Zhang, Yizhou Zhao, Ruijie Zheng,
  and Yuke Zhu.
\newblock Gr00t n1: An open foundation model for generalist humanoid robots,
  2025.
\newblock URL \url{https://arxiv.org/abs/2503.14734}.

\bibitem[Wan et~al.(2025)Wan, Wang, Ai, Wen, Mao, Xie, Chen, Yu, Zhao, Yang,
  Zeng, Wang, Zhang, Zhou, Wang, Chen, Zhu, Zhao, Yan, Huang, Feng, Zhang, Li,
  Wu, Chu, Feng, Zhang, Sun, Fang, Wang, Gui, Weng, Shen, Lin, Wang, Wang,
  Zhou, Wang, Shen, Yu, Shi, Huang, Xu, Kou, Lv, Li, Liu, Wang, Zhang, Huang,
  Li, Wu, Liu, Pan, Zheng, Hong, Shi, Feng, Jiang, Han, Wu, and
  Liu]{wan2025wanopenadvancedlargescale}
Team Wan, Ang Wang, Baole Ai, Bin Wen, Chaojie Mao, Chen-Wei Xie, Di~Chen,
  Feiwu Yu, Haiming Zhao, Jianxiao Yang, Jianyuan Zeng, Jiayu Wang, Jingfeng
  Zhang, Jingren Zhou, Jinkai Wang, Jixuan Chen, Kai Zhu, Kang Zhao, Keyu Yan,
  Lianghua Huang, Mengyang Feng, Ningyi Zhang, Pandeng Li, Pingyu Wu, Ruihang
  Chu, Ruili Feng, Shiwei Zhang, Siyang Sun, Tao Fang, Tianxing Wang, Tianyi
  Gui, Tingyu Weng, Tong Shen, Wei Lin, Wei Wang, Wei Wang, Wenmeng Zhou, Wente
  Wang, Wenting Shen, Wenyuan Yu, Xianzhong Shi, Xiaoming Huang, Xin Xu, Yan
  Kou, Yangyu Lv, Yifei Li, Yijing Liu, Yiming Wang, Yingya Zhang, Yitong
  Huang, Yong Li, You Wu, Yu~Liu, Yulin Pan, Yun Zheng, Yuntao Hong, Yupeng
  Shi, Yutong Feng, Zeyinzi Jiang, Zhen Han, Zhi-Fan Wu, and Ziyu Liu.
\newblock Wan: Open and advanced large-scale video generative models, 2025.
\newblock URL \url{https://arxiv.org/abs/2503.20314}.

\bibitem[Wang et~al.(2026)Wang, Li, Guan, Ye, Xie, Liu, Chen, Liang, Zhang, Hu,
  Huang, Lin, Lin, Liu, Bai, Zhou, Zhang, Yuan, Zhou, Yin, Wang, Huang, Lei,
  Peng, Chen, Zheng, Fan, Zhuang, Zhou, Li, Chen, Zhang, Liu, Sun, Chen, Li,
  Lü, Yang, Yu, and Chen]{wang2026qwenvlaunifyingvisionlanguageactionmodeling}
Qiuyue Wang, Mingsheng Li, Jian Guan, Jinhui Ye, Sicheng Xie, Yitao Liu, Junhao
  Chen, Zhixuan Liang, Jie Zhang, Xintong Hu, Xuhong Huang, Pei Lin, Junyang
  Lin, Dayiheng Liu, Shuai Bai, Jingren Zhou, Jiazhao Zhang, Haoqi Yuan, Gengze
  Zhou, Hang Yin, Ye~Wang, Yiyang Huang, Zixing Lei, Wujian Peng, Delin Chen,
  Yingming Zheng, Jingyang Fan, Xianwei Zhuang, Xin Zhou, Haoyang Li, Anzhe
  Chen, Tong Zhang, Xuejing Liu, Yuchong Sun, Ruizhe Chen, Zhaohai Li, Chenxu
  Lü, Zhibo Yang, Tao Yu, and Xionghui Chen.
\newblock Qwen-vla: Unifying vision-language-action modeling across tasks,
  environments, and robot embodiments, 2026.
\newblock URL \url{https://arxiv.org/abs/2605.30280}.

\bibitem[Xue et~al.(2025)Xue, Ren, Chen, Zhang, Fang, Gu, Xu, and
  Lu]{xue2025reactivediffusionpolicyslowfast}
Han Xue, Jieji Ren, Wendi Chen, Gu~Zhang, Yuan Fang, Guoying Gu, Huazhe Xu, and
  Cewu Lu.
\newblock Reactive diffusion policy: Slow-fast visual-tactile policy learning
  for contact-rich manipulation, 2025.
\newblock URL \url{https://arxiv.org/abs/2503.02881}.

\bibitem[Xue et~al.(2026)Xue, Rigo, Huang, Shen, Xu, Colonnese, and
  Memar]{xue2026tube}
Teng Xue, Alberto Rigo, Bingjian Huang, Jiayi Shen, Zhengtong Xu, Nick
  Colonnese, and Amirhossein~H Memar.
\newblock Tube diffusion policy: Reactive visual-tactile policy learning for
  contact-rich manipulation.
\newblock \emph{arXiv preprint arXiv:2604.23609}, 2026.

\bibitem[Yang et~al.(2025)Yang, Li, Yang, Zhang, Hui, Zheng, Yu, Gao, Huang,
  Lv, Zheng, Liu, Zhou, Huang, Hu, Ge, Wei, Lin, Tang, Yang, Tu, Zhang, Yang,
  Yang, Zhou, Zhou, Lin, Dang, Bao, Yang, Yu, Deng, Li, Xue, Li, Zhang, Wang,
  Zhu, Men, Gao, Liu, Luo, Li, Tang, Yin, Ren, Wang, Zhang, Ren, Fan, Su,
  Zhang, Zhang, Wan, Liu, Wang, Cui, Zhang, Zhou, and
  Qiu]{yang2025qwen3technicalreport}
An~Yang, Anfeng Li, Baosong Yang, Beichen Zhang, Binyuan Hui, Bo~Zheng, Bowen
  Yu, Chang Gao, Chengen Huang, Chenxu Lv, Chujie Zheng, Dayiheng Liu, Fan
  Zhou, Fei Huang, Feng Hu, Hao Ge, Haoran Wei, Huan Lin, Jialong Tang, Jian
  Yang, Jianhong Tu, Jianwei Zhang, Jianxin Yang, Jiaxi Yang, Jing Zhou,
  Jingren Zhou, Junyang Lin, Kai Dang, Keqin Bao, Kexin Yang, Le~Yu, Lianghao
  Deng, Mei Li, Mingfeng Xue, Mingze Li, Pei Zhang, Peng Wang, Qin Zhu, Rui
  Men, Ruize Gao, Shixuan Liu, Shuang Luo, Tianhao Li, Tianyi Tang, Wenbiao
  Yin, Xingzhang Ren, Xinyu Wang, Xinyu Zhang, Xuancheng Ren, Yang Fan, Yang
  Su, Yichang Zhang, Yinger Zhang, Yu~Wan, Yuqiong Liu, Zekun Wang, Zeyu Cui,
  Zhenru Zhang, Zhipeng Zhou, and Zihan Qiu.
\newblock Qwen3 technical report, 2025.
\newblock URL \url{https://arxiv.org/abs/2505.09388}.

\bibitem[Yang et~al.(2022)Yang, Haase-Schütz, Leonardi,
  et~al.]{yang2022pressurevision}
Patrick~Grady Yang, Christian Haase-Schütz, Marcel Leonardi, et~al.
\newblock Pressurevision: Estimating hand pressure from a single rgb image.
\newblock In \emph{European Conference on Computer Vision (ECCV)}, 2022.

\bibitem[Yuan et~al.(2026{\natexlab{a}})Yuan, Zhang, Zhou, Chen, Wang, Liu,
  Niu, Wang, Zhang, Zhang, Hu, Gong, Xing, Wen, Lu, Zhang, and
  Gao]{yuan2026ftp1generalistfoundationtactile}
Chengbo Yuan, Zicheng Zhang, Mingjie Zhou, Wendi Chen, Yi~Wang, Zhuoyang Liu,
  Dantong Niu, Shuo Wang, Hui Zhang, Wenkang Zhang, Yingdong Hu, Yuanqing Gong,
  Wanli Xing, Chuan Wen, Cewu Lu, Kaifeng Zhang, and Yang Gao.
\newblock Ftp-1: A generalist foundation tactile policy across tactile sensors
  for contact-rich manipulation, 2026{\natexlab{a}}.
\newblock URL \url{https://arxiv.org/abs/2606.13102}.

\bibitem[Yuan et~al.(2026{\natexlab{b}})Yuan, Yi, Zhang, Chen, Mo, Yin, Li,
  Zeng, Wen, Lu, Driggs-Campbell, and
  Lourentzou]{yuan2026vtamvideotactileactionmodelscomplex}
Haoran Yuan, Weigang Yi, Zhenyu Zhang, Wendi Chen, Yuchen Mo, Jiashi Yin,
  Xinzhuo Li, Xiangyu Zeng, Chuan Wen, Cewu Lu, Katherine Driggs-Campbell, and
  Ismini Lourentzou.
\newblock Vtam: Video-tactile-action models for complex physical interaction
  beyond vlas, 2026{\natexlab{b}}.
\newblock URL \url{https://arxiv.org/abs/2603.23481}.

\bibitem[Yuan et~al.(2017)Yuan, Dong, and Adelson]{yuan2017gelsight}
Wenzhen Yuan, Siyuan Dong, and Edward~H Adelson.
\newblock Gelsight: High-resolution robot tactile sensors for estimating
  geometry and force.
\newblock \emph{Sensors}, 17\penalty0 (12):\penalty0 2762, 2017.

\bibitem[Zang et~al.(2026)Zang, Zheng, Nie, Zheng, Tian, Gu, Gao, Wang, Yan,
  and Ding]{zang2026tacforesightforceguidedtactileworld}
Yujie Zang, Yuhang Zheng, Xian Nie, Yupeng Zheng, Shuai Tian, Songen Gu, Chen
  Gao, Zining Wang, Shuicheng Yan, and Wenchao Ding.
\newblock Tacforesight: Force-guided tactile world model for contact-rich
  manipulation, 2026.
\newblock URL \url{https://arxiv.org/abs/2606.11184}.

\bibitem[Zhang et~al.(2025)Zhang, Xu, Yang, Yue, Lin, ang Gao, Wang, and
  Zhao]{zhang2025tavlaelucidatingdesignspace}
Zongzheng Zhang, Haobo Xu, Zhuo Yang, Chenghao Yue, Zehao Lin, Huan ang Gao,
  Ziwei Wang, and Hao Zhao.
\newblock Ta-vla: Elucidating the design space of torque-aware
  vision-language-action models, 2025.
\newblock URL \url{https://arxiv.org/abs/2509.07962}.

\bibitem[Zhao et~al.(2023)Zhao, Kumar, Levine, and
  Finn]{zhao2023learningfinegrainedbimanualmanipulation}
Tony~Z. Zhao, Vikash Kumar, Sergey Levine, and Chelsea Finn.
\newblock Learning fine-grained bimanual manipulation with low-cost hardware,
  2023.
\newblock URL \url{https://arxiv.org/abs/2304.13705}.

\bibitem[Zhao et~al.(2025)Zhao, Haldar, Cui, Pinto, and
  Bhirangi]{zhao2025touchbeginsvisionends}
Zifan Zhao, Siddhant Haldar, Jinda Cui, Lerrel Pinto, and Raunaq Bhirangi.
\newblock Touch begins where vision ends: Generalizable policies for
  contact-rich manipulation, 2025.
\newblock URL \url{https://arxiv.org/abs/2506.13762}.

\bibitem[Zheng et~al.(2026{\natexlab{a}})Zheng, Niu, Xie, Wang, Xu, Jiang,
  Castañeda, Hu, Tan, Fu, Darrell, Huang, Zhu, Xu, and
  Fan]{zheng2026egoscalescalingdexterousmanipulation}
Ruijie Zheng, Dantong Niu, Yuqi Xie, Jing Wang, Mengda Xu, Yunfan Jiang,
  Fernando Castañeda, Fengyuan Hu, You~Liang Tan, Letian Fu, Trevor Darrell,
  Furong Huang, Yuke Zhu, Danfei Xu, and Linxi Fan.
\newblock Egoscale: Scaling dexterous manipulation with diverse egocentric
  human data, 2026{\natexlab{a}}.
\newblock URL \url{https://arxiv.org/abs/2602.16710}.

\bibitem[Zheng et~al.(2026{\natexlab{b}})Zheng, Gu, Li, Zheng, Zang, Tian, Li,
  Hao, Gao, Liu, Li, Chen, Yan, and
  Ding]{zheng2026omnivtavisuotactileworldmodeling}
Yuhang Zheng, Songen Gu, Weize Li, Yupeng Zheng, Yujie Zang, Shuai Tian, Xiang
  Li, Ce~Hao, Chen Gao, Si~Liu, Haoran Li, Yilun Chen, Shuicheng Yan, and
  Wenchao Ding.
\newblock Omnivta: Visuo-tactile world modeling for contact-rich robotic
  manipulation, 2026{\natexlab{b}}.
\newblock URL \url{https://arxiv.org/abs/2603.19201}.

\bibitem[Zhou et~al.(2026)Zhou, Gao, Hong, Liu, Zhang, Dai, Zhen, Lyu, Wu, Mao,
  Wang, Jiang, Ding, and Yang]{zhou2026touchanythingdatasetframeworkbimanual}
Jianyi Zhou, Ziteng Gao, Feiyang Hong, Zirui Liu, Guannan Zhang, Weisheng Dai,
  Ruichen Zhen, Chuqiao Lyu, Haotian Wu, Yinian Mao, Xushi Wang, Yuxiang Jiang,
  Wenbo Ding, and Shuo Yang.
\newblock Touchanything: A dataset and framework for bimanual tactile
  estimation from egocentric video, 2026.
\newblock URL \url{https://arxiv.org/abs/2605.13083}.

\bibitem[Zhou et~al.(2025)Zhou, Chen, Chen, Chen, Zhao, Jin, Ren, and
  Luo]{zhou2025act2goalworldmodelgeneral}
Pengfei Zhou, Liliang Chen, Shengcong Chen, Di~Chen, Wenzhi Zhao, Rongjun Jin,
  Guanghui Ren, and Jianlan Luo.
\newblock Act2goal: From world model to general goal-conditioned policy, 2025.
\newblock URL \url{https://arxiv.org/abs/2512.23541}.

\end{thebibliography}

\clearpage

\beginappendix
\section{Related Work}
\noindent\textbf{Vision-language-action robot foundation policies.}
Recent robot foundation policies have rapidly scaled vision-language-action models for general manipulation through large-scale robot data, vision-language pretraining, diffusion or flow-matching action decoders, and cross-embodiment training~\cite{intelligence2025pi05visionlanguageactionmodelopenworld,nvidia2025gr00tn1openfoundation,liu2025rdt1bdiffusionfoundationmodel,wang2026qwenvlaunifyingvisionlanguageactionmodeling}.
While these models improve semantic grounding and language-conditioned action generation, they remain primarily driven by visual observations and action-level supervision.
In contact-rich dexterous manipulation, key physical states such as slip, grasp stability, contact force, and insertion progress are difficult to infer from vision alone.
TouchWorld extends this line of work by introducing tactile-conditioned action generation and high-frequency tactile refinement into a hierarchical VLA framework.

\noindent\textbf{Tactile representation learning and tactile manipulation policies.}
Tactile sensing provides direct measurements of local contact states, including pressure, deformation, slip, and hand-object interaction, which are difficult to infer reliably from RGB observations alone.
High-resolution tactile sensors such as GelSight~\cite{yuan2017gelsight} and DIGIT~\cite{lambeta2020digit} have enabled dense contact perception, while prior work has studied visuo-tactile prediction~\cite{li2019visgel}, visual pressure estimation~\cite{yang2022pressurevision,grady2024pressurevision++}, and tactile estimation from egocentric human interaction~\cite{zhou2026touchanythingdatasetframeworkbimanual}.
Building on these tactile representations, recent tactile manipulation policies incorporate touch into robot action models for contact-rich manipulation.
FTP-1 scales tactile policy learning across heterogeneous tactile sensors and contact-rich tasks~\cite{yuan2026ftp1generalistfoundationtactile}, while T-Rex studies tactile-reactive dexterous manipulation by exploiting high-frequency tactile signals for reactive control~\cite{niu2026trextactilereactivedexterousmanipulation}.
These works demonstrate that tactile observations are valuable not only for perception but also for contact-aware action generation.
However, tactile observations are often fused into a monolithic policy as an additional modality, or used primarily as reactive cues within the action-generation loop.
TouchWorld instead assigns touch two complementary roles: a predictive signal for visual-tactile goal generation and a fast feedback signal for high-frequency residual refinement.
This separation allows the visuo-tactile policy to preserve semantic and geometric progress while the tactile refinement layer performs online contact adaptation.

\noindent\textbf{Hierarchical, predictive, and reactive manipulation policies.}
Long-horizon manipulation often benefits from separating high-level task structure from low-level control.
Prior work has explored action chunking and diffusion-based visuomotor policies to generate temporally coherent short-horizon action sequences rather than single-step commands~\cite{zhao2023learningfinegrainedbimanualmanipulation,chi2025diffusion,black2025realtimeexecutionactionchunking}.
Hierarchical robot policies further use intermediate representations such as subtasks, paths, or visual goals to bridge language-level reasoning and local execution~\cite{li2025hamsterhierarchicalactionmodels}.
Predictive policies and robot world models extend this idea by forecasting future visual states, goal trajectories, or latent dynamics to guide goal-conditioned action generation~\cite{zhou2025act2goalworldmodelgeneral,zang2026tacforesightforceguidedtactileworld}.
In parallel, reactive tactile policies and residual controllers use recent tactile or force feedback to correct local execution errors during contact-rich manipulation~\cite{xue2025reactivediffusionpolicyslowfast,zhao2025touchbeginsvisionends}.
However, these directions are often studied separately: hierarchical and predictive policies mainly rely on language or visual goals, while reactive tactile policies focus on local contact adaptation without explicit long-horizon semantic structure.
TouchWorld combines these ideas in a unified tactile VLA framework by coupling executable subtask planning, predictive visual-tactile goals, nominal action-chunk generation, and online tactile residual refinement.

\section{Implementation Details}

\noindent\textbf{High-Level Planning Layer.}
The High-Level Planning Layer contains the Subtask Planner and the Tactile World Model.
We instantiate the Subtask Planner from Qwen3-VL-4B-Instruct and adapt it with supervised LoRA fine-tuning.
The Subtask Planner takes the task instruction, the current camera observation, and a compact memory of recent subtasks and execution states, and predicts the executable subtask used by the downstream policy.
The High-Level Planning Layer is updated at a fixed slow semantic rate in our experiments.
To make the Subtask Planner robust to stale or repeated histories, the SFT set includes teacher labels, oversampled rare phases, and noisy-memory variants.
The resulting Subtask Planner dataset contains 128,866 records sampled at this semantic update rate.
We use LoRA rank 16, alpha 32, dropout 0.05, a learning rate of $10^{-4}$, cosine decay with 0.1 warmup ratio, bfloat16 training, and 20 epochs.
At inference time, only the selected subtask is passed to the manipulation policy; intermediate reasoning is used only for logging and analysis.

\noindent\textbf{Visuo-Tactile Goal-Conditioned Policy.}
The nominal manipulation policy is implemented as a visuo-tactile diffusion Transformer.
It receives multi-view RGB observations, the current tactile observation rendered as an image, the predicted goal grid from the Tactile World Model when available, proprioceptive state, and a language prompt that concatenates the original task and the current subtask.
In this VLA branch, tactile input is represented uniformly as image-form context rather than by separate tactile encoders for different sensor types.
The model encodes visual, tactile-image, and language inputs as context tokens and denoises action tokens with a Transformer action expert.
On the Wuji platform, the policy predicts a 120-dimensional action vector over a 32-step action horizon.
The action vector consists of two 48-dimensional arm-hand action representations, a 9-dimensional head action, and 15 reserved dimensions.
Training uses the teleoperation action targets and the normalization statistics computed from the zarr training set.
We train the nominal policy for 30,000 optimization steps with global batch size 32, bfloat16 precision, AdamW, gradient clipping at 1.0, and a cosine learning-rate schedule with 1,000 warmup steps, peak learning rate $2.5\times10^{-5}$, and final learning rate $2.5\times10^{-6}$.

\noindent\textbf{Tactile World Model.}
The Tactile World Model is fine-tuned from Wan2.2-TI2V-5B and predicts short-horizon visual-tactile subgoal observations conditioned on the current multimodal context and selected subtask.
Each training sample is formatted as a 2-by-2 observation grid containing external RGB views and tactile-image observations, together with the task and subtask text.
The Tactile World Model uses large-scale human interaction video data and 10 hours of robot demonstration data.
At 30 FPS, the robot demonstrations correspond to approximately 1.08 million robot frames.
Robot training samples use 17-frame clips at $384 \times 224$ resolution.
The world model is adapted with LoRA on the DiT attention and feed-forward projections, using target modules $\{q,k,v,o,\mathrm{ffn}.0,\mathrm{ffn}.2\}$, rank 64, learning rate $10^{-4}$, and 50 epochs.
During deployment, the High-Level Planning Layer is still updated at the slow semantic rate, but the Tactile World Model is refreshed only when the high-level memory indicates a meaningful subtask or task-state change.
If the current phase is unchanged, the system reuses the previous predicted goal, which avoids repeatedly invoking the world model during stable execution phases.

\noindent\textbf{Tactile-Conditioned Refinement Policy.}
The refinement layer is implemented as TRT, a compact Tactile Residual Transformer over recent tactile history, proprioceptive state, sliding nominal-action lookahead windows, and VLA context tokens.
Unlike the VLA branch, this layer uses structured tactile histories and modality-specific tactile tokenization for different tactile signal types.
The implementation uses a short causal tactile window that captures recent contact changes.
TRT takes a sliding nominal-action lookahead window with $W=16$ and predicts a residual window over the same horizon.
The nominal VLA still produces the full 120-dimensional action, but the residual layer is trained only on a 58-dimensional tactile-sensitive action subspace covering the two wrist pose blocks and selected hand joints.
The predicted residual is scattered back into the 120-dimensional action vector, while the remaining action dimensions are kept from the nominal VLA output.
The residual Transformer uses $d_{\mathrm{model}}=512$, eight layers, eight attention heads, a $W$-step residual query set, and a small residual regularization weight of $10^{-4}$.
In residual training, the trained nominal VLA is attached to the refinement layer and kept frozen; the residual actor and tactile feedback encoder are optimized with masked MSE between the corrected action window and the demonstration action window.
We use AdamW with learning rate $10^{-4}$, weight decay $10^{-4}$, and the same action normalization as the nominal policy.

\noindent\textbf{Deployment schedule.}
The full system uses a multi-rate execution schedule.
The High-Level Planning Layer runs at a slow semantic rate.
The Tactile World Model is called only when the Subtask Planner changes the subtask or contact-relevant phase.
The visuo-tactile policy generates nominal action chunks with horizon $H=32$.
Within each chunk, TRT refines sliding nominal-action lookahead windows with $W=16$ at stride $C=4$ (e.g., offsets $\{0,4,8,12\}$) and commits the first $C$ corrected actions before refreshing the residual prediction.
For example, TRT conditions on nominal actions $[0,15]$ and commits corrected actions $[0,3]$, then conditions on $[4,19]$ and commits $[4,7]$.
The realized feedback rate depends on the robot control frequency, sensing pipeline, and commit interval $C$.
If the Subtask Planner or Tactile World Model is unavailable, the system falls back to the original task prompt and disables predicted-goal conditioning, while the visuo-tactile policy and tactile refinement layer remain executable.

\end{CJK*}
\end{document}